\documentclass[conference]{IEEEtran}
\usepackage{cite}
\usepackage{amsmath,amssymb,amsfonts}
\usepackage{algorithmic}
\usepackage{graphicx}
\usepackage{textcomp}
\usepackage{xcolor}
\usepackage{soul}
\def\BibTeX{{\rm B\kern-.05em{\sc i\kern-.025em b}\kern-.08em
    T\kern-.1667em\lower.7ex\hbox{E}\kern-.125emX}}
\begin{document}

\title{Multi-objective tuning for torque PD controllers of cobots
}

\author{\IEEEauthorblockN{Diego Navarro-Cabrera}
  \IEEEauthorblockA{diegonavaca@ugr.es}
\and
\IEEEauthorblockN{Niceto R. Luque}
  \IEEEauthorblockA{nluque@ugr.es}
\and
\IEEEauthorblockN{Eduardo Ros}
  \IEEEauthorblockA{eros@ugr.es}
}
\maketitle

\begin{abstract}
  Collaborative robotics is a new and challenging field in the realm of motion control and human-robot interaction. The safety measures needed for a reliable interaction between the robot and its environment hinder the use of classical control methods, pushing researchers to try new techniques such as machine learning (ML). In this context, reinforcement learning has been adopted as the primary way to create intelligent controllers for collaborative robots, however supervised learning shows great promise in the hope of developing data-driven model based ML controllers in a faster and safer way. In this work we study several aspects of the methodology needed to create a dataset to be used to learn the dynamics of a robot. For this we tune several PD controllers to several trajectories, using a multi-objective genetic algorithm (GA) which takes into account not only their accuracy, but also their safety. We demonstrate the need to tune the controllers individually to each trajectory and empirically explore the best population size for the GA and how the speed of the trajectory affects the tuning and the dynamics of the robot.
\end{abstract}


\section{Introduction}
Collaborative robotics is an emerging field that studies the creation and development of robots designed for a safe human-machine interaction i.e. human-robot collaboration. The motion control of these cobotic systems is a complex problem since it incorporates both active safety measures, such as torque control that aims to minimize the force applied by the joints, and passive measures, like the integration of elastic elements that provide a higher level of compliance in case of an impact with humans or objects in the environment. These measures hinder the calculation of the analytical dynamic model of the cobot, which prevents the use of classical torque-based control algorithms that rely on widely used rigid simple models. Furthermore, position-based control is not well suited for human-robot interaction (HRI) as the commanded motion can carry significant levels of inertia, posing a risk to human safety.\\

To overcome the reliance on an analytical definition of system dynamics in traditional control theory, machine learning (ML) is being profusely used \cite{MLRobotics1}. ML offers promising control solutions for operating model-free dynamic systems, enabling accurate and safe task performance. Among various learning types, reinforcement learning emerges as the most prevalent due to its capability for generalization and data capture through practice \cite{MLRobotics2}. However, this learning approach does come with certain drawbacks for real systems, including a lengthy learning period and an exploration stage that can pose risks to both the robot and its environment \cite{MLRobotics2}.\\

As a result, in this work, we focus on studying the methodology required to create a database that enables the data-driven learning of a cobot's dynamic model, rather than calculating it analytically \cite{DynamicIdentification}. Building upon the previous discussion, our main goal is to generate a dataset that facilitates the study and development of supervised learning models, so that they can be used for avoiding risks during the learning stages with reinforcement learning or other adaptive control alternatives. This approach takes advantage of optimized position control scheme for gathering data.\\ 

The database we propose captures the relationship between the reached position and velocity of the cobot and the corresponding applied torque values. Depending on the direction of this relationship (reached position to applied torque values or vice versa), the database can serve as either an inverse dynamic model or a forward dynamic model of the cobot. This database is obtained by executing a representative set of trajectories with the cobot operating in torque control, guided by a proportional-derivative (PD) controller. The PD controller is adjusted using a multi-objective GA that optimizes movement precision and torque values to ensure safety. The extracted data from this process will be used to train the subsequent ML controller, providing optimal torque sequences for the cobot to accurately perform the desired trajectories, akin to accurate position control, while minimizing torque requirements.\\

The PD torque control requires precise adjustment of the PD parameters for each target trajectory. Each data sequence of torque value-reached position, obtained from individual PD adjustments, is generated specifically to train a subsequent ML controller. This ML controller will be able to generalize the control action and adapt it to various types of trajectories \cite{DynamicModel}.

\section{Related work}
The PD control architecture is widely used in robotic manipulators due to its simplicity \cite{ControlReview}. This technique involves adjusting only two parameters per robotic joint and provides accurate control for simple tasks within a limited range of motion.\\ 

PD adjustment using GA is widely used in the industry, leading to a wide range of proposed GA techniques \cite{PIDReview}. While most of these works focus on single-objective GA techniques, in our collaborative robot approach the goal is not only to maximize controller accuracy but also to ensure HRI safety by minimizing torque values, PD adjustment requires the use of multi-objective GA. An example of such an GA is the NSGA-II \cite{NSGA}, which enables the optimization of multiple control goals  simultaneously.\\

The adjustment of PD controllers using multi-objective GA has been previously addressed by \cite{MOPID}, where a PID (proportional-integral-derivative) controller was tuned using NSGA-II. In \cite{Cuckoo} multi-objective cuckoo search algorithm (MOCSA) is used for the same problem but no comparison between algorithms is provided so we cannot say whether MOCSA is more appropriate than NSGA-II for this problem. \cite{MOPID-II} also uses NSGA-II and compares it with a variation of the same algorithm which uses decision maker preference information to reduce the decision parameters' search space. All of these results, while promising, were only tested in simulation with a relatively simplistic planar two-degree-of-freedom (d.o.f.) robot arm model. In this work, our aim is to validate the effectiveness of the NSGA-II solution using the more complex Kuka iiwa LBR robot arm, equipped with 7 d.o.f. and flexible joints \cite{Kuka}.\\

Despite the proven usefulness of learned dynamic cobot models \cite{DynamicModel}, to the best of our knowledge, there is currently no publicly available dataset that captures the relationship between torque values and motion of a cobot in a manner suitable for learning its dynamic model. Therefore, the objective of this work is to present and discuss the methodology used to collect the necessary data required for learning a dynamic cobot model

\section{Proposed solutions}
To ensure a balance between optimal torque utilization and accuracy in the collected data, we will utilize a custom-tuned PD controller. As mentioned earlier, PD adjustment can result in highly accurate torque-based control for specific trajectories. However, as we will demonstrate later, the accuracy diminishes significantly when performing dissimilar trajectories located far from the PD working point.\\

For the PD adjustment, we propose the use  of a multi-objective GA to jointly optimize accuracy and safety, specifically maximizing accuracy while minimizing the torque values involved. To achieve this, we incorporate two objectives within the objective functions. The first objective assigns weight to the accuracy error, measured as the mean Euclidean distance between the end effector and the desired Cartesian coordinates. Meanwhile, the second objective assigns weight to the torque values applied throughout the trajectory. The torque values are calculated using Eq. (\ref{ft}), where $U$ represents the vector of commanded torques, $T$ denotes the number of steps in a trajectory, and $u_i$ corresponds to the torque applied at time $i$.\\

\begin{equation}
f_t(U)=\frac{1}{T}\sum_{i=1}^{T}(u_i-u_{i-1})^2\label{ft}
\end{equation}

\begin{figure}[htbp]
\centerline{\includegraphics[width=0.9\columnwidth]{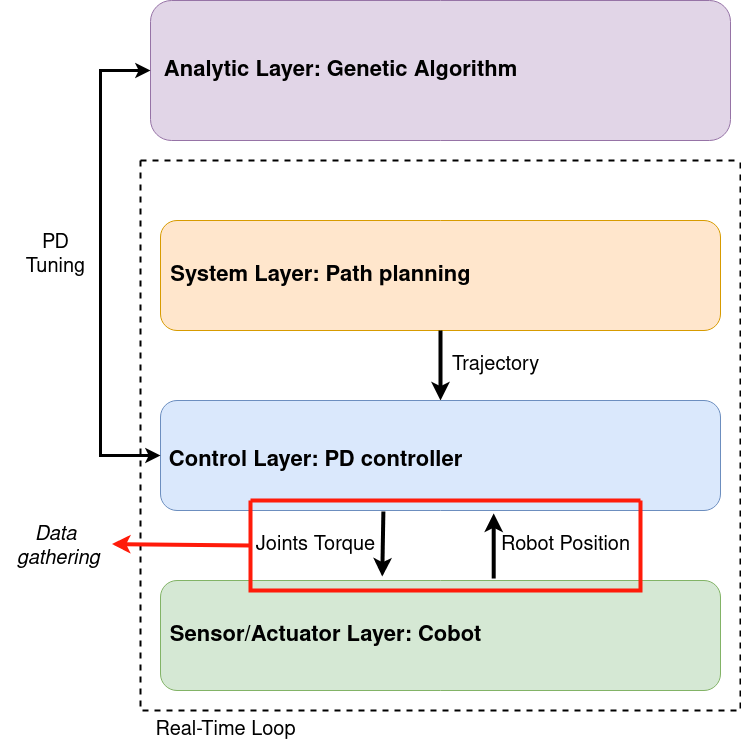}}
\caption{Architecture of the proposed system. General framework adapted from IMOCO4.E \cite{ImocoFramework}. First the system layer sends the desired setpoint to the controller, then the control layer sends torque commands to the cobot and finally the sensor layer returns the position and velocity of each joint. The extracted data is saved into a file used asynchronously by the analytic layer to update the PD controller gains.}
\label{architecture} 
\end{figure}

Our methodology divides the data collection process into four main layers, as shown in Figure \ref{architecture}:
\begin{itemize}
\item Sensor/Actuator layer: This layer comprises the sensors and actuators used by the cobot. It receives instructions from the controller and provides data on the joint states.
\item Control layer: The PD controller is located at this layer and receives information regarding the next desired set-point as well as the parameters (Kp and Kd) to be used. It sends corresponding torque commands.
\item System layer: This layer sends data about the desired trajectory to the control layer.
\item Analytic layer: The GA in this layer is used to adjust the PD controller gains based on  the system performance.
\end{itemize}

System, control and actuator/sensor layers work on a real-time loop at 500Hz. In this period of time the system layer sends the trajectory to the control layer which sends the torque command to the cobot and receives the updated sensor data. The torque, position and velocity of each joint is registered in an array and written to a file once the trajectory is finished. Once the the data file is created, the analytic layer reads it to evaluate performance and comunicates asynchronously with the control layer to update the PD gains.\\

This division facilitates the scalability of our methodology by separating the analytic and system layers from the control and sensor layers. Furthermore, it allows for the parallel utilization of multiple cobots with the same trajectory.\\

\begin{figure}[htbp]
  \includegraphics[width=0.45\columnwidth]{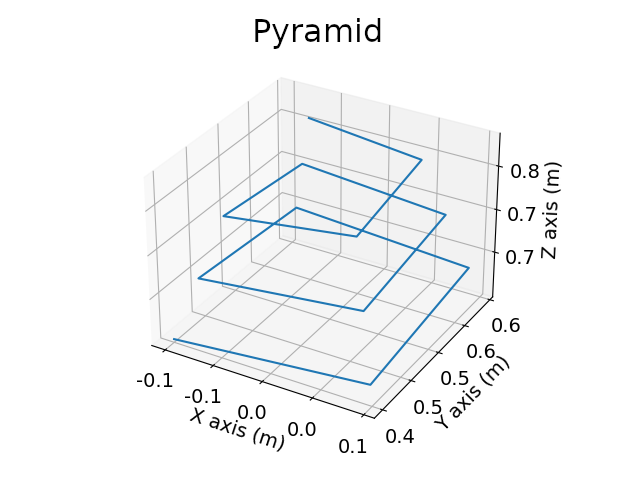}
  \includegraphics[width=0.45\columnwidth]{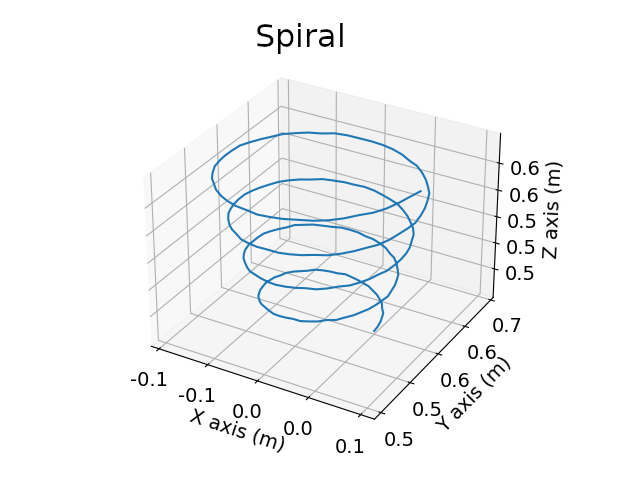}
  \centerline{\includegraphics[width=0.45\columnwidth]{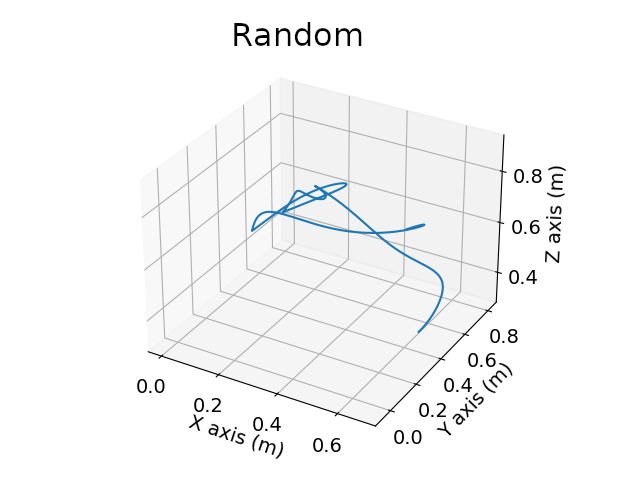}}
\caption{Examples of useful trajectories for data gathering.}
\label{trajectories}
\end{figure}

Regarding the trajectories included in the dataset, and following the findings in \cite{DynamicModel}, we incorporate spiral and random trajectories. These trajectory types generate meaningful data sets while avoiding excessive data size, making them suitable for effective training of the ML controller. Additionally, we introduce pyramid-like trajectories that combine linear movements with sharp turns. These trajectories aim to better teach a ML controller how to function when working with high acceleration and velocity gradients, resulting in larger inertia values. Fig. \ref{trajectories} depicts some examples of the trajectory dataset.\\

\begin{figure}[htbp]
\centerline{\includegraphics[width=0.7\columnwidth]{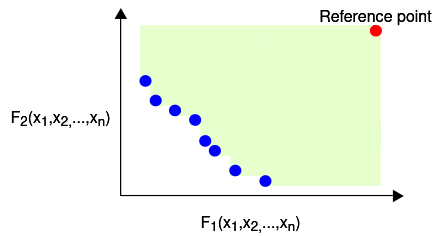}}
\caption{Diagram depicting the calculation of the hypervolume indicator. This metric is obtained measuring the area between the pareto front and a reference point.}
\label{hypervolume}
\end{figure}

Finally, to compare the GA solutions for the PD parameters, we utilize the hypervolume indicator metric \cite{Hypervolume}. This metric takes a reference point (e.g. the maximun values between all the controllers tested). It then calculates the area between the Pareto front and this reference point. A visual example of this metric can be seen in Figure \ref{hypervolume}\\

\section{Progess to date}
The results presented in this work are obtained from a simulated environment (and the application and validation in a non-simulated environment are left for future work). For our robotic simulation platform, we use ROS2 (Robot Operating System) \cite{ROS2}, and Gazebo \cite{Gazebo} as our dynamic simulator. The close integration of Gazebo with ROS makes it a suitable choice for the performed study.\\

Each experiment in this section is repeated 5 times accounting for the stochastic nature of the GA. This number of trials balances computation time (over a couple of weeks) and reliability of the data and conclusions. Box plots are used to represent the locality and spread of the results. These experiments are conducted to demonstrate the feasibility of the proposed methodology.\\

\begin{figure}[htbp]
  \includegraphics[width=\columnwidth]{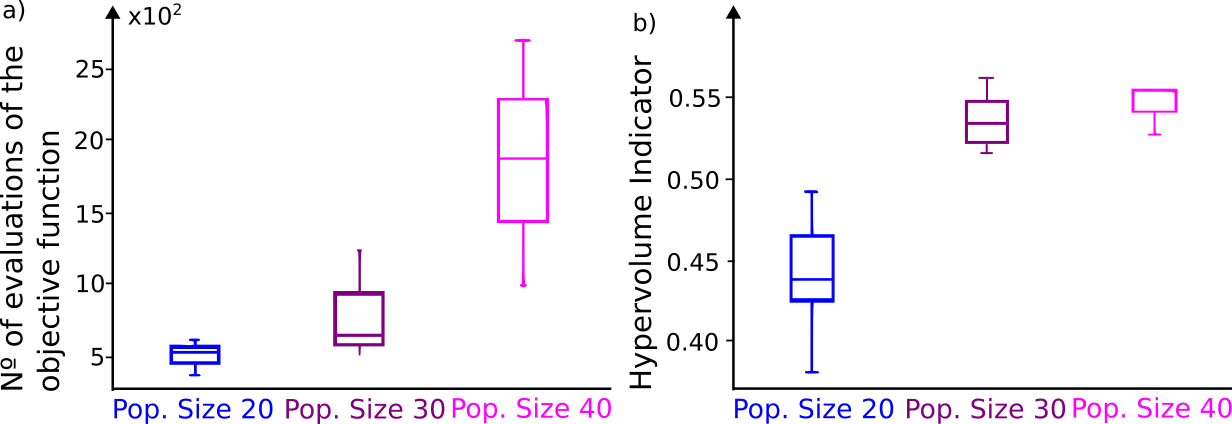}
\caption{Comparison of the number of evaluations needed for convergence (a) and the hypervolume of the obtained pareto front (b) based on population size.}
\label{Pop size}
\end{figure}

In one set of experiments, various population sizes are compared to find the optimal balance between accuracy and computation time. Fig. \ref{Pop size} depicts that the algorithm (NSGA-II) achieves the best results with a population size of around 30 individuals. Increasing the population size carries similar results, but with significantly longer execution times.

\begin{figure}[htbp]
  \includegraphics[width=\columnwidth]{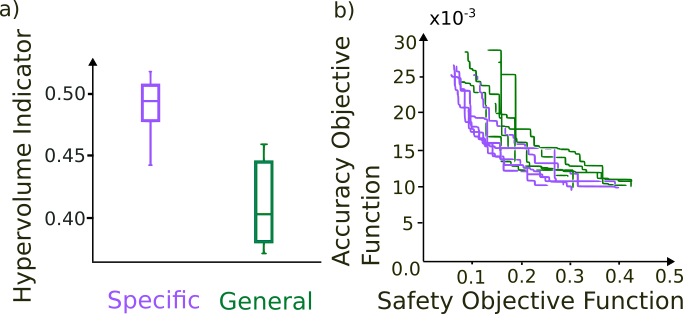} 
\caption{Hypervolume indicator (a) and pareto fronts (b) obtained using a general controller and one tuned to the specific trajectory. Results obtained using a piramid trajectory similar to the one in Fig. \ref{trajectories}.}
\label{GS}
\end{figure}

Once the GA is configured, we compare the accuracy achieved by a generic track controller and a specific track controller. The generic controller is adjusted to perform globally on all trajectories in the dataset, while the specific controller is tuned for a single trajectory. Fig. \ref{GS} demonstrates that the specific controller outperforms the generic controller, not only in terms of precision but also in minimizing the applied torque values. This indicates that while it is possible to achieve high accuracy (at least in simulation) by overloading the joint motors, achieving smooth and safe movements requires a well-tuned specific PD controller. Since the goal of the PD optimization is to be able to perform different movements with different optimal accuracy/torque profiles, it is key to use a specific optimized controllers for each trajectory. Then, all the data gathered from the different trajectories (and specific controllers) will be added to the database. This specific optimization stage is required, because during the trajectory execution stage (gathering the dataset) it is captured both the robot dynamics but also the properties of the controller used. Thus optimizing specific controllers leads to a richer database in terms of accuracy and torque trade-off (Figure \ref{GS}).\\

Finally, the speed of the trajectory is one of the key factors that significantly impacts the dynamics of a cobot. Thus, we investigate the extent to which the speed of the trajectory influences the PD adjustment and determine the optimal speed at which the trajectories in our dataset shall be executed.\\

\begin{figure}[htbp]
\centerline{\includegraphics[width=\columnwidth]{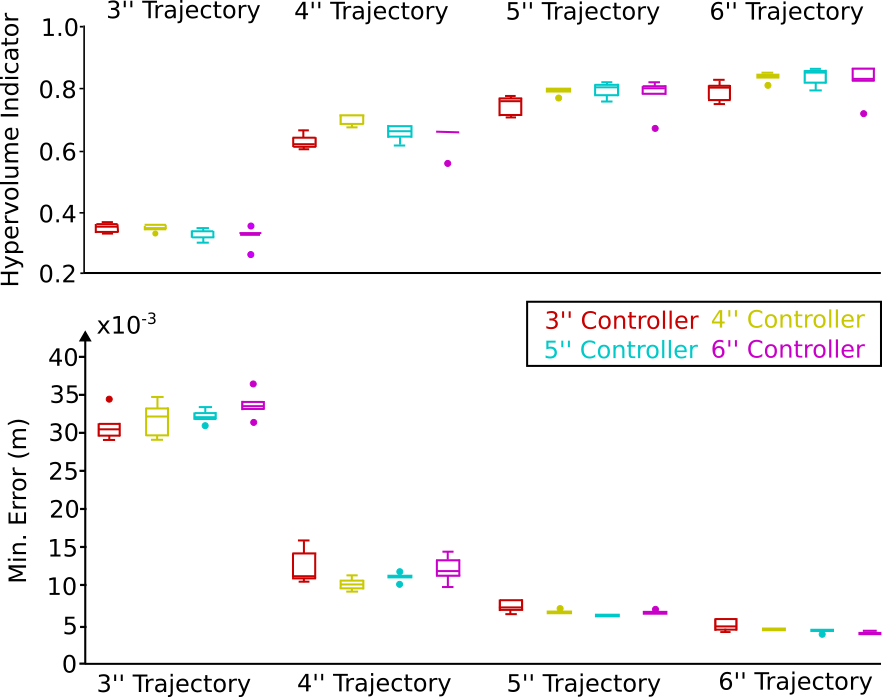}}
\caption{Study on the effect of trajectory speed to the accuracy of a set of controllers. On the top the hypervolume indicator is measured, while on the bottom the mean cartesian error of the controller with highest accuracy is represented.}
\label{time}
\end{figure}

To accomplish this, we create multiple variations of the same target trajectory (a spiral), each with a different duration ranging from 3 to 6 seconds. This range was selected on the consideration that faster trajectories are not achievable, and slower trajectories would exhibit negligible differences in their dynamics. As the duration of the trajectory increases, the motion commands required for the cobot to track it become slower. Next, we adjust a set of PD controllers for each individual trajectory and assess their performance on the other trajectories. The results of this study are illustrated in Figure \ref{time}, where $X'' controller$ represents a set of controllers that were specifically adjusted using a trajectory of $X$ seconds.\\

From these results, two conclusions can be drawn. Firstly, the speed of the trajectory has a notable impact on the accuracy of the control, with accuracy rapidly decreasing at higher speeds. The accuracy stabilizes at around 5 seconds, making it the optimal duration for this trajectory as it strikes the best balance between execution time and controller accuracy.\\

Secondly, although there is a slight drop in performance when transferring a controller from one trajectory to another, the differences between sets of controllers are relatively small. This suggests that at this regime, the speed of the trajectory does not significantly affect the PD adjustment  but rather the data gathered.

\section{Conclusions and future research}
The work presented here was focused on defining a methodology to create a dataset from which most ML solutions were able to capture the dynamics model of a cobot. To collect optimal tuples of torque-position/velocity data, we applied multi-objective GAs to finely adjust PDs that controlled the torque of a cobot throughout its working space, maximizing accuracy and minimizing torque values.\\

In future work, we aim to apply this methodology to a non-simulated cobot platform covering the sim2real gap and demonstrating  how  the main concepts indicated in this work also apply for real robots. Although the specific trajectories and optimized controllers may differ when addressing the GA, the presented work and results provide valuable insights. It is important to note that the intensive optimization effort presented in this work cannot be directly performed on a robotic platform due to the potential risk it poses to the robot integrity.

\section*{Acknowledgment}

This study was supported by the EU with the IMOCOe4.0 [EU H2020RIA-101007311] project and by Spanish national funding [PCI2021-121925]. This study was also supported by SPIKEAGE [PID2020-113422GA-I00] by the Spanish Ministry of Science and Innovation MCIN/AEI/10.13039/501100011033, awarded to NRL; DLROB [TED2021-131294B-I00] funded by MCIN/AEI/10.13039/501100011033 and by the European Union NextGenerationEU/PRTR, awarded to NRL; MUSCLEBOT [CNS2022-135243] funded by MCIN/AEI/10.13039/501100011033 and by the European Union NextGenerationEU/PRTR, awarded to NRL.

\end{document}